# Cross-Institutional Dental EHR Entity Extraction via Generative AI and Synthetic Notes

Yao-Shun Chuang[1], Chun-Teh Lee[2], Guo-Hao Lin[3], Ryan Brandon[4], Xiaoqian Jiang[1], Muhammad F. Walji[1], Oluwabunmi Tokede[5, 6]

[1] McWilliams School of Biomedical Informatics, The University of Texas Health Science Center at Houston, 7000 Fannin Street, Houston, TX 77030
[2] Department of Periodontics and Dental Hygiene, The University of Texas Health Science Center at Houston School of Dentistry, 7500 Cambridge St, Houston, TX 77054
[3] Postgraduate Periodontics Program, School of Dentistry, University of California San Francisco, 707 Parnassus Avenue, D-3015, San Francisco, CA 94143, USA
[4] Willamette Dental Group and Skourtes Institute, 5935 SE Alexander St, Hillsboro, OR 97123
[5] Oral Healthcare Quality and Safety, Diagnostic and Biomedical Sciences, The University of Texas Health Science Center at Houston School of Dentistry, 7500 Cambridge St, Houston, TX 77054
[6] Diagnostic and Biomedical Sciences, The University of Texas Health Science Center at Houston School of Dentistry, 7500 Cambridge St, Houston, TX 77054

**Abstract**

*Background:*

While most healthcare providers now use electronic health records (EHRs) to document clinical care, many still treat them as digital versions of paper records. As a result, documentation often remains unstructured, with free-text entries in progress notes. This limits the potential for secondary use and analysis, as machine learning and data analysis algorithms are more effective with structured data.

*Objective:*

This study aims to use advanced artificial intelligence (AI) and natural language processing (NLP) techniques to improve diagnostic information extraction from clinical notes in a periodontal use case. By automating this process, the study seeks to reduce missing data in dental records and minimize the need for extensive manual annotation, a longstanding barrier to widespread NLP deployment in dental data extraction.

*Methods:*

This research utilizes Large Language Models (LLMs), specifically GPT-4, to generate synthetic medical notes for fine-tuning a RoBERTa model. This model was trained to better interpret and process dental language, with particular attention to periodontal diagnoses. Model performance was evaluated by manually reviewing 360 clinical notes randomly selected from each of the participating site's dataset.

*Results:*

The results demonstrated high accuracy of periodontal diagnosis data extraction, with the Site 1 and 2 achieving a weighted average score of 0.97-0.98. This performance held for all dimensions of periodontal diagnosis in terms of stage, grade and extent.

*Conclusion:*

The study highlights the potential transformative impact of AI and NLP on healthcare research. Most clinical documentation (40-80%) is free text. Scaling our method could enhance clinical data reuse.

**Introduction**

The vast amount of health data generated in the United States has the potential to unlock critical insights into disease understanding, quality improvement, and cost reduction in healthcare[1,2]. Electronic dental records (EDRs), which are digital collections of patient oral healthcare events and observations, are now ubiquitous and essential for healthcare delivery, operations, and research in dentistry[3–5]. EDR data is often categorized as structured and unstructured. Structured EDR data includes standardized diagnoses, medications, and laboratory values in fixed numerical or categorical fields; however, challenges such as missing, incomplete, and inconsistent data are prevalent. Unstructured data, on the other hand, consists of free-form text written by healthcare providers, such as clinical notes and discharge summaries. Dental care providers frequently document detailed findings, diagnoses, treatment plans, and prognostic factors in this free-text format[6]. This unstructured data, which comprises a significant portion of total health records, is readily accessible during patient care but presents substantial challenges for secondary analysis, often requiring time-consuming and costly manual review by domain experts to extract meaningful insights[7–9].

To effectively address this challenge, natural language processing (NLP) techniques offer a robust solution for extracting data from unstructured clinical entries. One fundamental NLP task in this context is Named Entity Recognition (NER), which involves identifying and extracting specific entities of interest—such as disease diagnoses, medication names, and laboratory test results—from clinical narratives[10]. Over the past decade, the focus of NLP approaches has largely centered on developing supervised, semi-supervised, and unsupervised NER systems[11]. However, with the rapid advancement in hardware and computational technologies, neural network-based NER systems have become increasingly prevalent, offering enhanced accuracy and scalability in recent years[12–14]. While numerous studies have shown exceptional performance in NER tasks, these models lack generalizability and require significant time to develop due to their dependency on extensive data[15,16].

Another major challenge for the widespread deployment of NLP in healthcare is the tedious and time-consuming process of manual annotation[17]. In order for NLP models to accurately interpret and analyze clinical text, large volumes of data must be annotated with precise labels, such as identifying medical conditions, medications, and patient outcomes. This task, often performed manually by experts, is not only labor-intensive but also prone to human error[18]. The sheer volume of clinical records, combined with the complexity and variability of medical language, makes this process even more burdensome. Consequently, the scalability of NLP applications in healthcare is hindered, as manual annotation becomes a bottleneck in training and validating machine learning models[17]. Without more efficient and automated approaches to data labeling, the potential of NLP to transform healthcare will be limited.

As a result, more recently the NLP techniques are being boosted by the advent of Large Language Models (LLMs) with deep learning-based architectures excelling in understanding and manipulating human language[19,20]. Trained on vast datasets, these models perform various language tasks with exceptional precision and are increasingly available as pre-trained models, facilitating easy integration into diverse applications without substantial computational resources or deep machine learning expertise. As NLP and LLMs continue to advance, their transformative impact is anticipated to expand, solidifying their role as essential elements of modern medical infrastructure[21–23].

This study aims to tackle the challenge of missing structured clinical data through a periodontal use case. In mitigating the challenges faced by previous attempts at deploying NLP techniques in dentistry, we deploy Generative Pre-trained Transformer 4 (GPT-4), to help enhance the efficiency of extracting diagnostic information related to periodontal diseases from dental clinical notes.

**Methods**

*Dataset*

Data for this study was sourced from an Electronic Dental Record (EDR) spanning January 1, 2021, to December 31, 2021 from two sites – UTHealth School of Dentistry (Site 1) and the University of California at San Francisco School of Dentistry (Site 2). The inclusion criteria required participants to have completed a thorough periodontal assessment, recording measurements such as pocket depth, clinical attachment loss, and the distance from the free gingival margin to the cemento-enamel junction (CEJ). Participants had to be at least 16 years old, possess a minimum of 10 natural teeth, and have a full mouth series of radiographs. These criteria are essential for precise periodontal diagnosis.

*Target information*

This study incorporated five dimensions of entities for diagnoses related to periodontal health. The entity of periodontal status encompasses three categories: periodontitis, gingivitis, and health, each associated with corresponding entity details specific to each category. For periodontitis, the entities assessed included stage, grade, and extent. The molar/incisor pattern extent was excluded from consideration due to its rarity. This approach aligns with the classification system established by the 2018 guidelines from the American Academy of Periodontology (AAP) and the European Federation of Periodontology (EFP)[24]. Specifically, diagnoses made prior to the implementation of the 2018 guidelines were excluded from the gold standard. In diagnosing gingivitis, the entities examined included the extent and the subtype (intact periodontium/reduced periodontium; past or stable periodontitis/ non-periodontitis); for gingival health, the focus was solely on the subtype. Details of the entities and their associated values are listed in Table 1.

**Table 1.** The details of entities.

| Entity | Values | Entity | Values |
|---|---|---|---|
| **Periodontal status** | Periodontitis | Stage | I, II, III, IV |
| | | Grade | A, B, C |
| | | Extent | Localized, Generalized |
| | Gingivitis | Subtype | Intact Periodontium, Reduced Periodontium; Past/Stable Periodontitis Non-Periodontitis |
| | | Extent | Localized, Generalized |
| | Health | Subtype | Intact Periodontium, Reduced Periodontium; Past/Stable Periodontitis, Non-Periodontitis |

*NLP building blocks*

The methodology of this study leveraged large language models (LLMs) to generate synthetic training data, enhancing the efficiency of developing a Named Entity Recognition (NER) model. To achieve this, the OpenAI GPT-4 model was integrated into a Python-based project via the OpenAI API to facilitate the generation of clinical notes. The process, illustrated in Figure 1, began with the extraction of target diagnosis notes, which served as structured templates to ensure adherence to clinical documentation standards. From Site 1, 15 clinical notes were randomly selected for each periodontal status category—periodontitis, gingivitis, and gingival health—identified by using a simple rule-based method, resulting in a total of 45 notes. Each of these template notes was then used to generate 10 synthetic clinical notes via the GPT-4 API, yielding a total of 450 synthetic notes. To ensure HIPAA compliance and protect patient privacy, all operations were conducted on UTHealth's HIPAA-compliant Azure OpenAI platform, maintaining full adherence to applicable privacy regulations.

In addition, the model was executed ten times per template to ensure diversity and data robustness. Prompts were meticulously designed with specific rules, components, and labeling instructions to align with domain-specific standards and contextual requirements in the dental field (Appendix A, Figure 1). The hyperparameters for temperature and top-p were both set to 1, while all other settings remained at their default values. This structured approach significantly improved the accuracy and efficiency of clinical note creation, demonstrating the potential of AI in automating complex text-based tasks within healthcare.

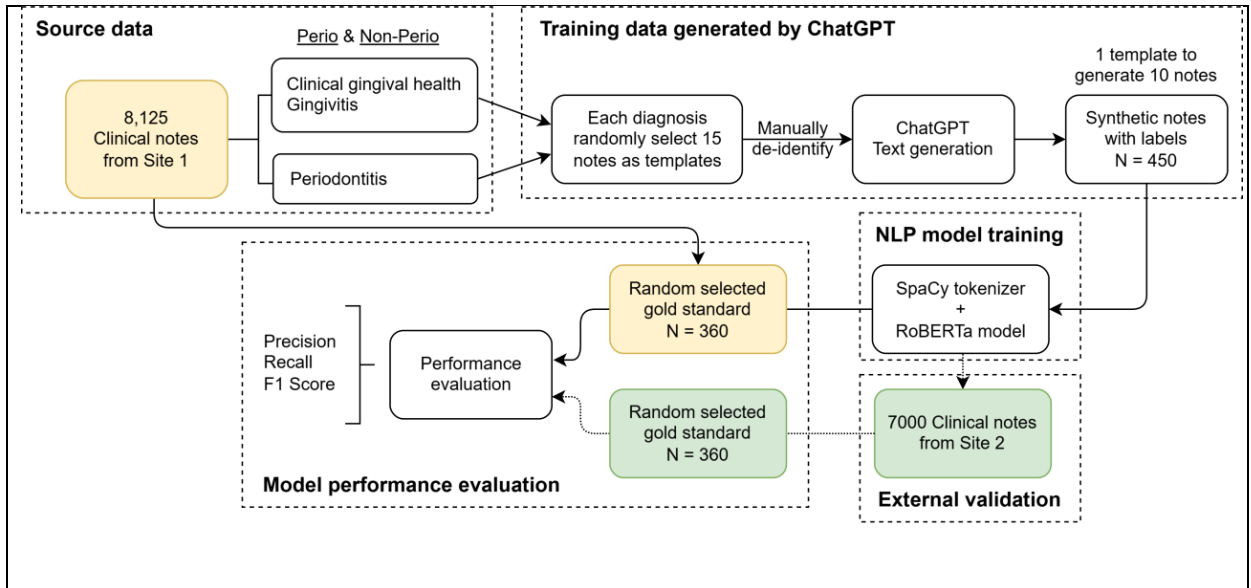

*Figure 1*: Flowchart of the Study.

*NER Model Setting*

Tokenization was performed using spaCy, a widely recognized NLP tool known for its non-destructive tokenization, which preserves the original structure of clinical notes, including whitespace and punctuation, ensuring data integrity. Its flexible architecture allows for customization of the NLP pipeline to meet specific needs. The dataset was split in an 8:1:1 ratio for training, validation, and testing, adhering to spaCy's training specifications. To enhance Named Entity Recognition (NER) performance, this study incorporated the RoBERTa base model, a robust pre-trained transformer. By leveraging spaCy's tokenization, the fine-tuned RoBERTa base model improved the identification and categorization of named entities within clinical notes. Additionally, RoBERTa's dynamic masked language modeling during pretraining mitigates over-memorization, making it particularly effective for tasks requiring precise information extraction[25,26].

*Improving Data Quality*

To enhance the quality of the data, we implemented additional steps both before and after processing. Prior to processing, we ensured that the labels and values extracted from the notes were accurate and suitable for training the model. Following processing, we made necessary corrections to ensure consistency, such as rectifying typographical errors and standardizing the reporting of stages and grades of conditions. In instances where data could not be generalized, we left those values blank to prevent inaccuracies.

In accordance with the 2018 AAP/EFP guidelines, each patient should only have one periodontal diagnosis. In cases where multiple diagnoses were present, we selected the most severe condition, prioritizing periodontitis over less severe conditions. For stages and grades, we retained the highest values, and for extent, we selected a generalized category when both generalized and localized conditions were indicated. This approach is consistent with standard clinical practices.

*Evaluating Model Performance*

To assess the performance of our fine-tuned model, we conducted a thorough evaluation using 360 clinical notes randomly selected from each of the two sites. These notes were first manually reviewed and annotated by domain experts to establish a gold standard. The NER model's predictions were then compared against these gold-standard annotations to evaluate its accuracy in identifying key diagnostic entities. This evaluation provided a comprehensive assessment of the model's effectiveness, ensuring both its robustness and potential for broader deployment across different institutional settings.

The NER model's performance was assessed using a confusion matrix. True positives (TP) are correct identifications of periodontal status, true negatives (TN) are agreements on the absence of the status, false positives (FP) are incorrect predictions, and false negatives (FN) are missed detections. Furthermore, three metrics evaluated the algorithm's effectiveness: Precision (P), Recall (R), and F1 Score. Precision measures the accuracy of true positives among positive predictions. Recall assesses the proportion of true positives among actual positive cases. The F1 Score is the harmonic mean of precision and recall. Due to variability and potential imbalance in periodontitis cases, both macro and weighted averages were used. The macro average computes the mean evaluation values across categories, while the weighted average accounts for the frequency of each category in the dataset.

**Results**

After applying the criteria, a total of 8,125 records from Site 1 and 7,000 records from Site 2 were identified. From each site, 360 clinical notes were then randomly selected to evaluate the NER model's performance in identifying target entities, specifically Periodontal Status, Stage, Grade, Extent, and Subtype. Although the ideal quantity for a gold standard set is typically around 10% of the total (approximately 812 and 700 notes from each dataset, respectively), given the relatively straightforward nature of the target entities without complex hierarchical structures, the number of gold standard annotations was gradually increased in increments of 30 notes. This iterative approach revealed that model performance stabilized at around 360 annotated notes (Figure 2). The performance data across the five dimensions demonstrated high accuracy, particularly in the Periodontal Status, Stage, and Grade categories (Figure 3). Both sites recorded a weighted average of 0.97 for Periodontal Status (Table 2), with both macro and weighted averages displaying minimal deviation, indicating consistent performance across classes. In the Stage and Grade categories, both sites achieved scores close to 0.97 for all evaluation metrics.

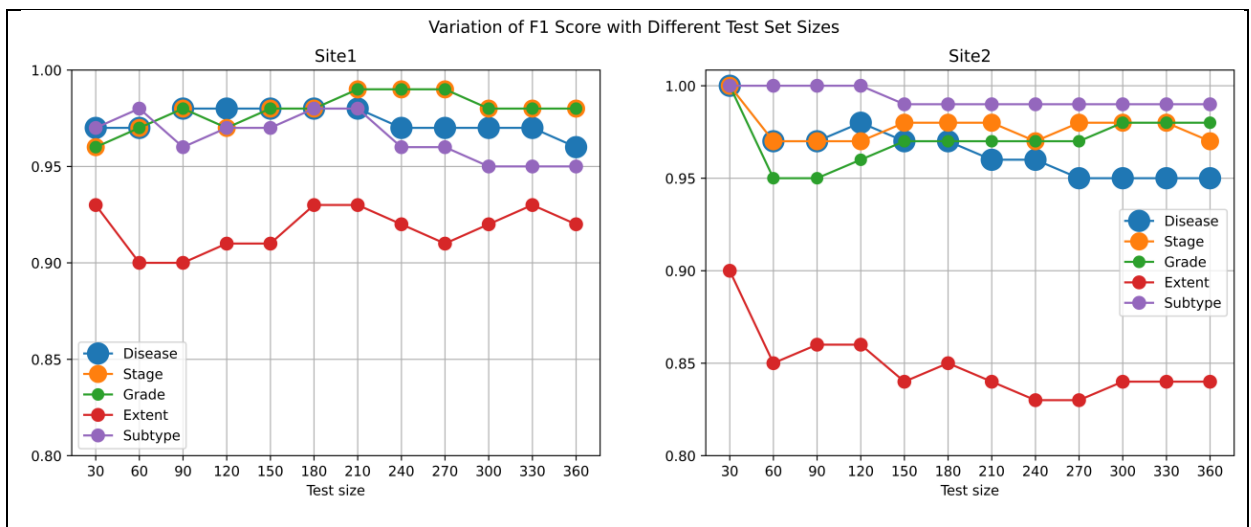

*Figure 2*: Variation of F1 Score with Different Test Set Sizes.

**Table 2.** The NER model performance comparison tested on the gold standard of the Sites 1 and 2.

|  |  | Precision | | Recall | | F1-score | |
|---|---|---|---|---|---|---|---|
|  | Average | Site 1 | Site 2 | Site 1 | Site 2 | Site 1 | Site 2 |
| **Periodontal status** | Macro | 0.96 | 0.92 | 0.96 | 0.96 | 0.96 | 0.94 |
|  | Weighted | 0.97 | 0.96 | 0.96 | 0.95 | 0.96 | 0.95 |
| **Stage** | Macro | 0.99 | 0.99 | 0.97 | 0.95 | 0.98 | 0.97 |
|  | Weighted | 0.98 | 0.98 | 0.98 | 0.97 | 0.98 | 0.97 |
| **Grade** | Macro | 0.99 | 0.99 | 0.96 | 0.95 | 0.97 | 0.97 |
|  | Weighted | 0.98 | 0.98 | 0.98 | 0.98 | 0.98 | 0.98 |
| **Extent** | Macro | 0.93 | 0.87 | 0.91 | 0.79 | 0.92 | 0.81 |
|  | Weighted | 0.93 | 0.86 | 0.92 | 0.85 | 0.92 | 0.84 |
| **Subtype** | Macro | 0.96 | 0.98 | 0.84 | 0.99 | 0.88 | 0.98 |
|  | Weighted | 0.95 | 0.99 | 0.95 | 0.99 | 0.95 | 0.99 |

For the Extent category, the model showed variability, with lower scores compared to other categories (Figure 3). At the Site 1, macro and weighted averages were 0.92-0.93 across all metrics (Table 2). The Site 2 had lower performance, with macro averages of 0.87 for precision, 0.79 for recall, and 0.81 for F1-score, and weighted averages of 0.86 for precision, 0.85 for recall, and 0.84 for F1-score (Table 2). In the Subtype category, the Site 2 generally outperformed the Site 1. The Site 1 had weighted averages of 0.95 for precision, recall, and F1-score. The Site 2 achieved near perfect performance with 0.99 in weighted averages (Table 2).

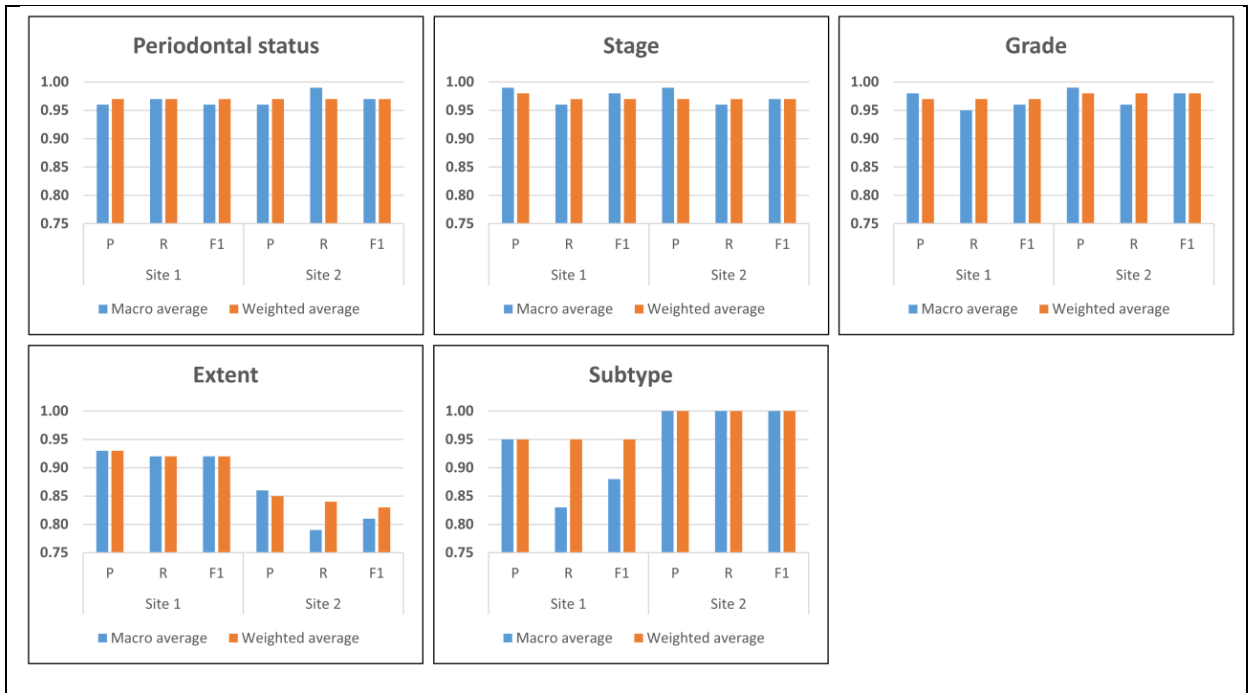

*Figure 3*: Bar Chart of NER Performance.

The confusion matrices were instrumental in evaluating the NER model's performance across various labels in the Sites 1 and 2 gold standard datasets (Figure 4). Specifically, analysis of the Extent entity within the Site 2 dataset revealed several classification errors, including missed cases, incorrect predictions between Localized and Generalized categories, and false positives. Additionally, the analysis highlighted a predominance of N/A cases in two labels of the Subtype entity, suggesting complexity in subcategory distinctions.

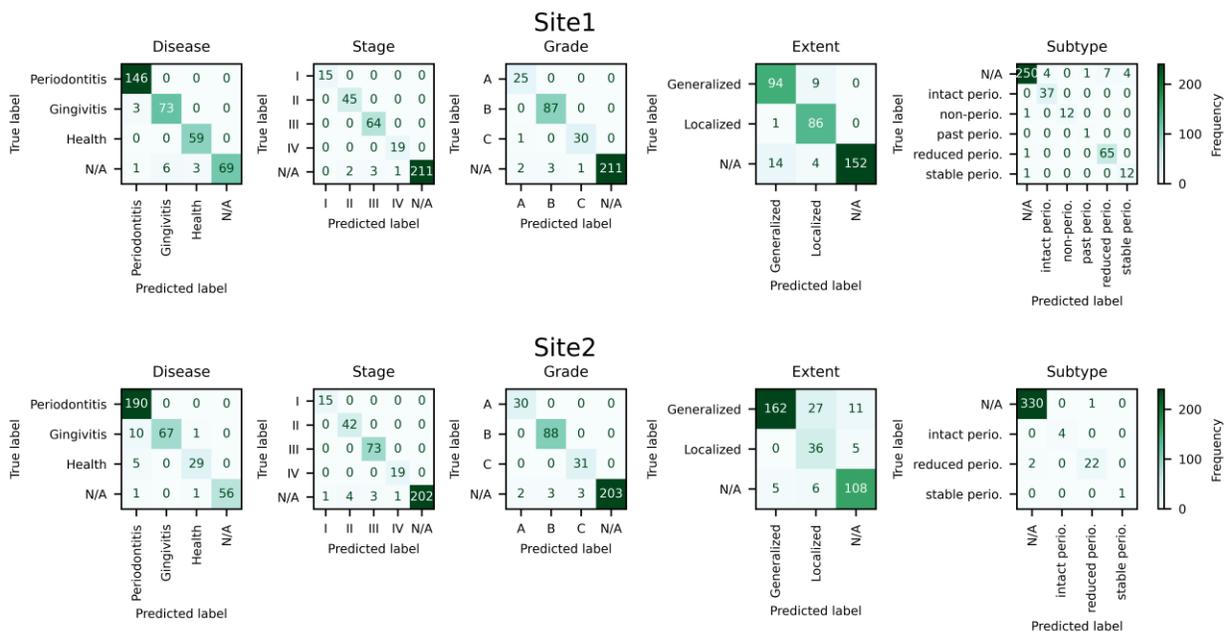

*Figure 4*: Confusion Matrix for Sites 1 and 2.

**Discussion**

This study tackled the significant clinical challenge of missing structured diagnoses in dental records by employing LLMs within an NLP framework. Our results demonstrate that a RoBERTa base NLP model, finetuned on synthetic data generated by large language models, accurately and efficiently identifies and extracts key diagnostic details from real clinical notes. The robust cross-site performance also indicates that this approach can generalize to varying clinical documentation styles, an essential feature for multi-institutional deployment. Secondary analysis of EHR data is becoming a critical tool for evidence generation. As such, a primary impetus for this work was to reduce the labor-intensive process of manual annotation usually associated with mining progress notes for valuable clinical information. While our results do not suggest the complete elimination of human oversight especially for complex or ambiguous notes, they do illustrate that the amount of expert-labeled data required to reach strong model performance can be significantly lowered. By generating 450 synthetic notes from 15 seed templates, we avoided having to manually annotate hundreds or thousands of real dental notes.

Additionally, the study revealed that synthetic data produced by LLMs effectively captures the essential features and variations of periodontal diagnoses, transcending specific institutional characteristics and highlighting the universal applicability and critical role of synthetic data in ensuring model effectiveness across diverse environments. The deployment of NLP techniques in dentistry is not novel, as many investigators have applied text-mining and machine-learning techniques for diagnosis extraction[5,6]. Some have used BERT or other transformer architectures to classify or extract periodontal conditions with solid accuracy. Our work extends this literature by innovatively leveraging a GPT-4–based data synthesis pipeline, which can be seen as complementary to previous approaches. Compared to those earlier studies that relied on traditional NLP tools or manually built NER models, which often struggled with data annotation[13,14,16,27], the use of synthetic training data generated by LLMs helped us to significantly streamline the data extraction process. Our models demonstrated generalizability, as evidenced by their success on external datasets (site 2), making them viable for resource-limited institutions. This scalability was key to expanding access to advanced AI tools in healthcare. The robust performance across datasets reflected a comprehensive training foundation, enabling adaptability to diverse dental terminologies and diagnostic styles. The local distillation process further refines the models by balancing the broad capabilities of LLMs with institution-specific needs. This approach enhances both specialization and versatility, improving NLP tools for critical tasks like identifying missing diagnoses while reducing the manual burden of data annotation. Nevertheless, it is important to recognize that this study was conducted at two academic institutions, both of which operate within academic primary and multispecialty dental care settings rather than general primary care environments. While the promising results suggest the potential for broader application, additional studies across a wider range of clinical settings would be valuable to further assess and confirm the generalizability of our approach.

In the qualitative analysis, the diverse and complex content presented a notable challenge, limiting the model's prediction accuracy and data processing capabilities, as identified through an in-depth examination of error cases. Most errors occurred in scenarios involving multiple related terms, typographical mistakes, formatting issues, and contextual entity extraction inaccuracies. In cases involving multiple related terms, significant errors arose when processing multiple "Extent" descriptors due to non-periodontal diagnosis descriptions or instances where multiple diagnoses were provided. For example, in the sentence, "D: Localized Periodontitis Stage I Grade A with Generalized Recession," the NLP system identified "Generalized" as the Extent. However, the correct Extent for periodontitis in this context should have been "Localized," resulting in erroneous entity extraction. Similarly, in cases with multiple diagnoses, such as "D: Localized Periodontitis Stage I Grade A and Generalized Periodontitis Stage II Grade B" or "D- Localized Periodontitis Stage I Grade A and Generalized Gingivitis," the guidelines specified in the methods section require extracting the most severe diagnosis. Although the model correctly extracted "Stage II," "Grade B," and "Generalized," it incorrectly assigned "Localized" as the corresponding Extent for "Stage II" and "Grade B." Addressing these errors would necessitate the development of an additional NLP model focused on Relation Extraction (RE) to resolve this issue; however, such development was beyond the scope of this study. This limitation was observed across both institutions, with a more pronounced impact in the Site 2 dataset, significantly contributing to the lower performance in identifying the Extent entity within that dataset. Another, though less frequent, error involved contextual entity extraction, as illustrated by the sentence, "Diagnosis: Stage III Grade B but to be confirmed with radiographs."

Another source of error arose from variations in formatting. The NER model implicitly analyzed and learned the relative positioning of entities based on the training data. In most reviewed cases, diagnostic sentences began with terms such as "D" or "Diagnosis." However, when a diagnosis appeared directly at the start of a sentence, such as "Generalized Stage 3 Grade B," the model typically predicted the Stage and Grade but omitted the Extent, in this case, "Generalized." Furthermore, informal diagnostic phrases—more frequently observed in the Site 2 dental notes, such as "Generalized III B" or "Stage 3 B"—highlighted a limitation in the synthetic training data generation. The template data and synthetic diagnostic sentences generated by LLMs did not include these direct diagnostic patterns. Although further fine-tuned models could potentially recognize this format of entities, the accuracy may remain inconsistent compared to standard-format diagnostic sentences.

Beyond error analysis, the NER model proposed in this study could facilitate the identification of both the completeness and the version of diagnoses, thereby aiding in the monitoring of diagnostic quality provided by dental care providers as well as in categorizing patient groups. During the case review, leveraging the reliability of the model's remarkable performance, a significant number of periodontal diagnoses made before the 2018 guidelines were identified in both institutions by verifying whether Stage and Grade were extracted. This ratio could be indicative of the care providers' awareness and serve as a metric to further evaluate their adherence to most recent diagnostic guidelines.

**Conclusions**
This study highlighted the novel deployment of LLMs in improving NER models for identifying and extracting periodontal diagnoses in clinical notes. The integration of synthetic training data streamlined the training process and enhanced model accuracy and adaptability. These locally distilled NER models reduced manual labor and improved model performance, especially with complex data formats and terminologies. Despite challenges with linguistic complexities and data formatting, the models' scalability and robustness indicate the potential for broader use in dental and medical diagnostics. Future enhancements could include integrating relation extraction techniques to improve entity recognition. Overall, this study provides a foundation for using advanced AI tools in healthcare, improving the interpretation of unstructured data, and enhancing patient care and healthcare management.

**Funding Statement**

This work was supported by the U.S. Department of Health and Human Services, National Institute of Dental and Craniofacial Research. Research Grant no.: 1R56DE034086 entitled: "FullMouth: Enhancing Dental Clinical Data and Reducing Disparities through Innovative ML Approaches."

**Competing Interests Statement**

No competing interests to declare.

**Contributorship Statement**

Yao-Shun Chuang (Conceptualization, Data curation, Formal analysis, Methodology, Validation, Visualization, Writing – original draft), Chun-Teh Lee (Formal analysis, Methodology, Writing – review & editing), Guo-Hao Lin Lee (Formal analysis, Methodology, Writing – review & editing), Ryan Brandon (Conceptualization, Resources, Formal analysis), Xiaoqian Jiang (Conceptualization, Project administration, Resources, Writing – review & editing), Muhammad Walji (Conceptualization, Project administration, Resources, Writing – review & editing), Bunmi Tokede (Conceptualization, Funding acquisition, Project administration, Resources, Writing – review & editing)

**Data Availability statement**

The data underlying this article cannot be shared publicly due to the presence of Protected Health Information (PHI) and the need to protect patient privacy in accordance with institutional and regulatory guidelines.

**Figure/Table Caption List**
Figure 1: Flowchart of the Study.
This figure presents a flowchart illustrating the study's methodology. The process includes the synthesis of training data, the fine-tuning of the NER model, and subsequent internal and external evaluations.

Figure 2: Variation of F1 Score with Different Test Set Sizes.
This figure presents the variation of F1 score with different sizes of test set for Site 1 and 2.

Figure 3: Bar Chart of NER Performance.
This figure presents a bar chart depicting the performance of the NER model on the gold standard dataset across five entities.

Figure 4: Confusion Matrix for Sites 1 and 2.
This figure presents the confusion matrix for Sites 1 and 2, comparing the gold standard annotations with the predictions made by the NER model.

Table 1: Details of Entities.
This table lists the types of entities and their corresponding values.

Table 2: Comparison of NER Model Performance.
This table presents a comparison of the NER model performance, tested on the gold standard datasets of Sites 1 and 2.